\documentclass[10pt,twocolumn,letterpaper]{article}

\usepackage[pagenumbers]{cvpr}  

\usepackage{graphicx}
\usepackage{capt-of}
\usepackage{xcolor}
\usepackage{amsmath,amssymb,amsfonts,dsfont,pifont,bm,bbm,mathrsfs,mathtools,nicefrac}
\usepackage{algorithm,algpseudocode,listings}
\usepackage{booktabs,multirow,adjustbox,diagbox,threeparttable}
\usepackage{bbding}
\usepackage{newfloat}
\usepackage{float}
\usepackage{makecell}
\definecolor{citeblue}{RGB}{48,111,186}
\usepackage[pagebackref,breaklinks,colorlinks,bookmarks=false]{hyperref}
\usepackage[capitalize]{cleveref}  
\crefname{section}{Sec.}{Secs.}
\Crefname{section}{Section}{Sections}
\crefname{table}{Tab.}{Tabs.}
\Crefname{table}{Table}{Tables}
\crefname{figure}{Fig.}{Figs.}
\Crefname{figure}{Figure}{Figures}
\crefname{equation}{Eq.}{Eqs.}
\Crefname{equation}{Equation}{Equations}
\newcommand{\tocite}[1]{\textcolor{red}{[TO CITE]}}
\newcommand{\method}{RiDDLE\xspace}

\def\cvprPaperID{1347}
\def\confName{CVPR}
\def\confYear{2023}

\begin{document}

\title{\method: Reversible and Diversified De-identification with Latent Encryptor}

\author{Dongze Li$^{1, 2}$, Wei Wang$^{2 *}$, Kang Zhao$^{3}$, Jing Dong$^{2}$, Tieniu Tan$^{2, 4}$\\
	$^{1}$ School of Artificial Intelligence, University of Chinese Academy of Sciences\\
	$^{2}$ Center for Research on Intelligent Perception and Computing, CASIA\\
 $^{3}$ Alibaba Group \ $^{4}$ Nanjing University\\
	\texttt{dongze.li@cripac.ia.ac.cn} \ \ \texttt{\{wwang, jdong, tnt\}@nlpr.ia.ac.cn}\thanks{ Corresponding author.} \\
 \texttt{zhaokang.zk@alibaba-inc.com}
	}

\maketitle

\begin{abstract}

This work presents \method, short for \textbf{R}evers\textbf{i}ble and \textbf{D}iversified \textbf{D}e-identification with \textbf{L}atent \textbf{E}ncryptor, to protect the identity information of people from being misused.
Built upon a pre-learned StyleGAN2 generator, \method manages to encrypt and decrypt the facial identity within the latent space.
The design of \method has three appealing properties.
First, the encryption process is cipher-guided and hence allows diverse anonymization using different passwords.
Second, the true identity can only be decrypted with the correct password, otherwise the system will produce another de-identified face to maintain the privacy.
Third, both encryption and decryption share an efficient implementation, benefiting from a carefully tailored lightweight encryptor.
Comparisons with existing alternatives confirm that our approach accomplishes the de-identification task with better quality, higher diversity, and stronger reversibility.
We further demonstrate the effectiveness of \method in anonymizing videos.
Code is available in \href{https://github.com/ldz666666/RiDDLE}{https://github.com/ldz666666/RiDDLE}.


\end{abstract}

\section{Introduction}

Recent advances in deep learning and computer vision technology bring convenience together with security concerns. Personal images shared on social media can be collected and abused by unauthorized software and malicious attackers, posing a threat to the privacy of individuals. 
Comparing with all other biometrics, face has unique importance because of its extensive application scenarios and abundant personal information. Face de-identification aims to hide the identity in the face image or video stream for privacy protection. Traditional de-identification methods such as blurring and mosaicing can effectively obfuscate the identity information, but the protected image is often severely damaged and loses its utility.

Current de-identification methods can generate a similar-looking person with a changed identity and have improved the image quality and utility by a large margin. However, many existing works~\cite{hukkelaas2019deepprivacy,gafni2019live,wu2018privacy} tend to generate faces with homogeneous appearances for different people, 
which is easy for an hacker to realize that these faces are anonymized.  In some cases, e.g. online conferences \cite{maximov2020ciagan}, the de-identified faces of participants should be different from each other.
For reliable identity protection, it is also necessary to consider variations in ethnicity, age, gender and other facial features.
Therefore, diversity is important to face identification. 
 Meanwhile, it is vital for the anonymous faces to keep superior image quality and utility. The former brings better photo-realism and stronger safety, while the latter makes it possible to perform identity-agnostic task on an anonymous face image in a privacy-preserving manner.  
\begin{figure}[t]
\centering
\includegraphics[width=0.91\linewidth]{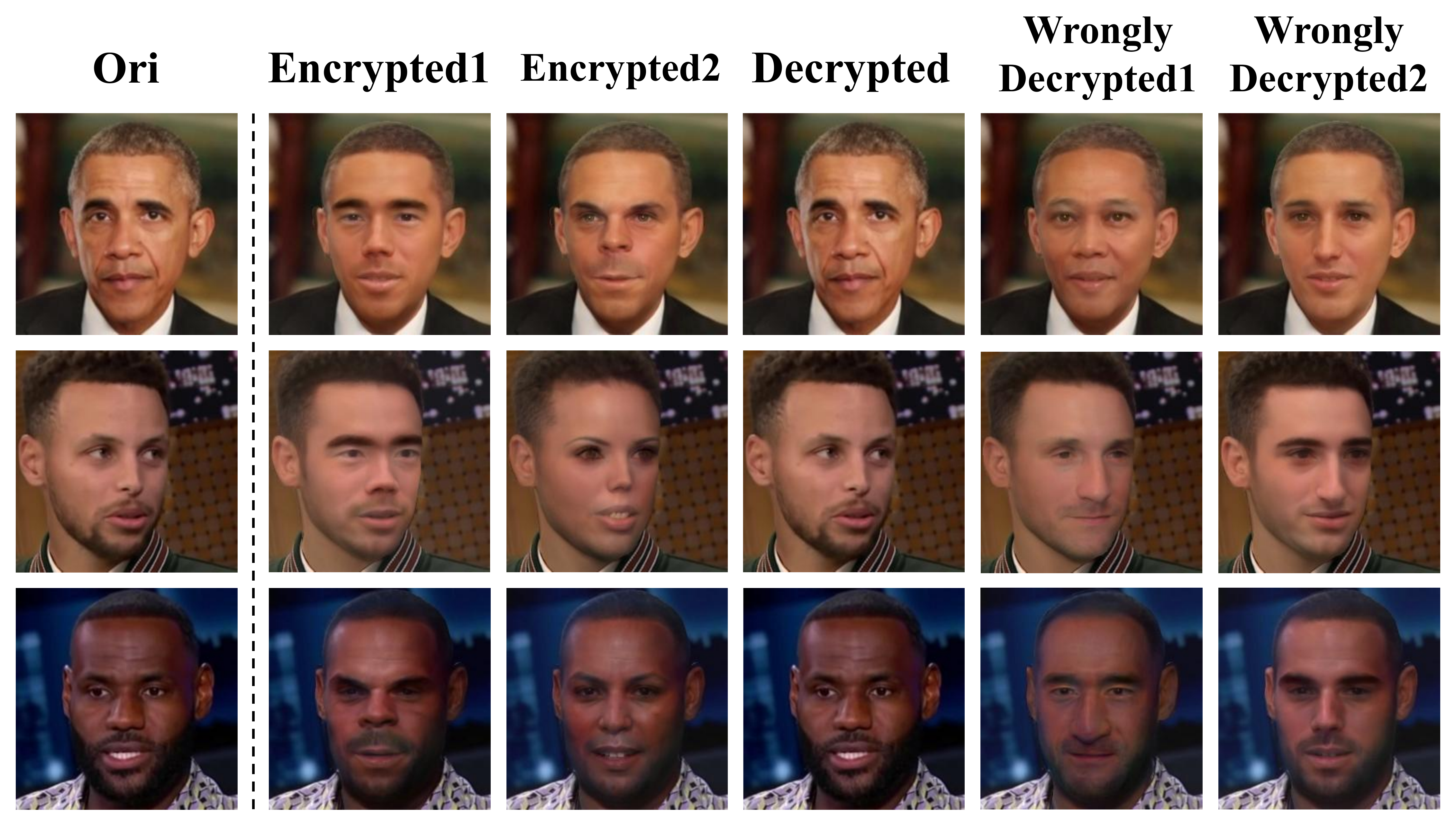} 
\caption{Encryption and decryption on images in the wild. The first column shows the original people. The second and third columns show different encrypted faces according to different passwords. The fourth column is the correctly decrypted faces, and the last two columns are the incorrectly decrypted faces. }
\label{fig1}
\end{figure}

Moreover, most of the previous works only pay attention to the process of de-identification, and neglect the importance of identity recovery.
Reversibility of the de-identification system is also crucial. For instance, family members are more willing to see the real face rather than the de-identified one and people may want to share data that only certain authorized parties can interpret.

Generally, a well-developed de-identification model should hold the following properties.
a) Maintaining high quality and the utility of the anonymous faces, as well as the identity independent attributes. 
b) Generating diversified virtual identities for safer privacy protection. 
c) Recovering the original faces when security conditions are satisfied.




To achieve the above goals, we propose \method, which is short for \textbf{R}evers\textbf{i}ble and \textbf{D}iversified \textbf{D}e-identification with \textbf{L}atent \textbf{E}ncryptor. The main features of our framework are shown below.

\textbf{Better Quality.} First, we project the face image onto the latent space of a strong generator, StyleGAN2\cite{karras2020analyzing}. Due to the decoupling characteristics of the face manifold, the identity independent attributes can be largely preserved. At the same time, high-quality virtual faces can be synthesized. De-identification and recovery results on images in the wild are shown in Figure \ref{fig1}.

\textbf{Higher Diversity.} After obtaining the inverted latent code, we perform encryption and decryption with a lightweight latent encryptor together with several randomly sampled passwords. In the encryption phase, each password is associated with a unique identity. Discrepancy between different anonymous faces is maximized, resulting in high diversity. 

\textbf{Stronger Reversibility.} In the decryption phase, when the password is correct, the true identity can be restored. Otherwise, a new de-identified face with photo-realism is returned. Different from opponents \cite{gu2020password} and \cite{cao2021personalized} which can achieve identity recovery to some extent, \method is free from manually designed encryption rules, does not need to be retrained for different passwords, and brings fewer visual artifacts. 

Another advantage of performing latent encryption is that when face datasets are unavailable due to privacy reasons, randomly sampled latents can be used to train the encryptor. We evaluate \method on both face image and video de-identification tasks. Sufficient experiments on public datasets and in the wild data have proven the superiority of \method over previous works.

\section{Related Work}
\subsection{Face De-identification}
Face de-identification aims to protect identity information from being recognized by unauthorized computer vision systems.
Classic de-identification methods, including blurring, mosaicing, masking, pixelation, and so on, can erase the identity thoroughly, but also cause serious detriment to the utility of the original images. 


\begin{figure*}[t]
\centering
\includegraphics[width=0.9\linewidth]{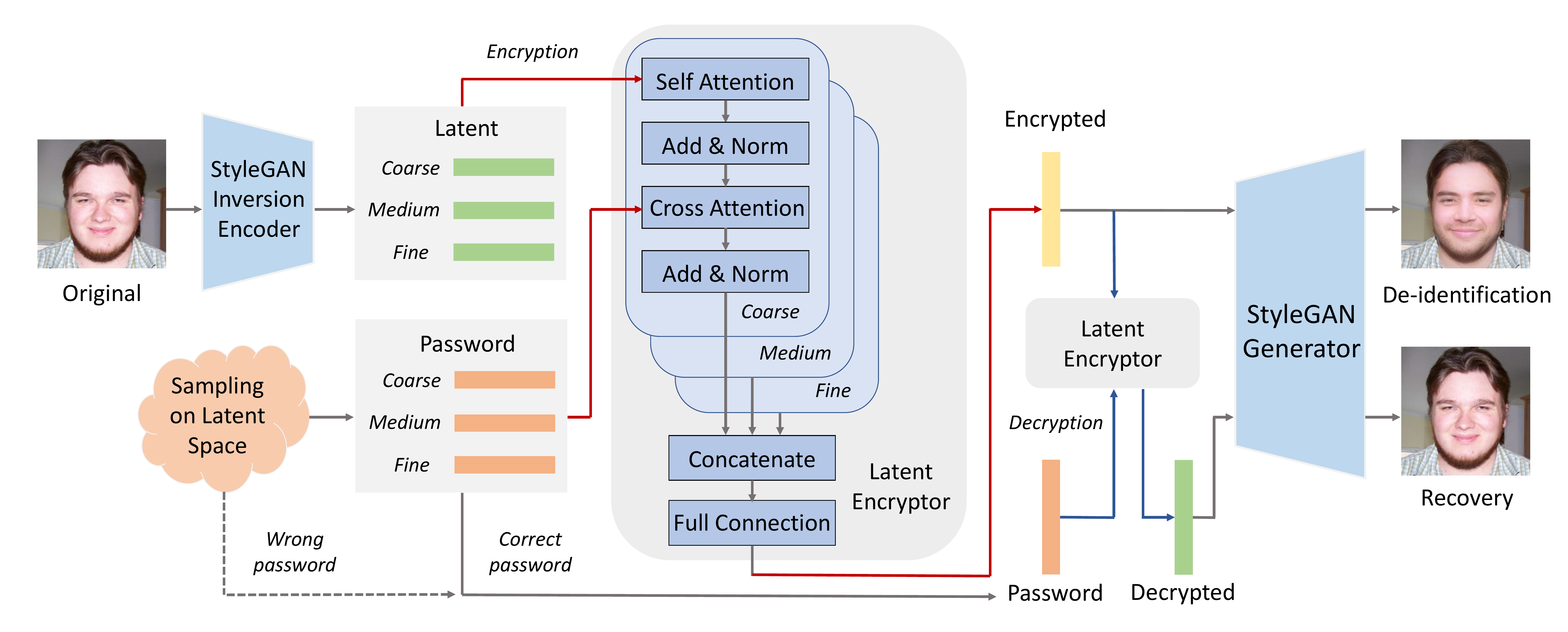} 
\caption{The overall pipeline of our method. We take the correct decryption process as an example, where the same password is used for both encryption and decryption.}
\label{fig2}
\end{figure*}

Recent de-identification methods focus on generating virtual faces to replace the original identity, while keeping their utility for downstream tasks. Some works \cite{hukkelaas2019deepprivacy,wu2018privacy,gafni2019live} realize de-identification by maximizing distance of identity embeddings with an identity dissimilarity loss term or multiple discriminators. \cite{gafni2019live} uses a multi-level perceptual loss to keep the low-level similarity while pushing away the higher. Other works \cite{sun2018natural,sun2018hybrid,ren2018learning,hukkelaas2019deepprivacy} tend to mask the facial area and synthesize a new face by inpainting. The works mentioned above can only map the original face to a single anonymized one and the de-identified faces often have similar appearance, and are usually considered unnatural. 
Another series of work uses predefined conditional labels to control the content of generated images \cite{maximov2020ciagan}, or decouples images into different representations and performs manipulation \cite{nitzan2020face,cao2021personalized}. 
Although various anonymized faces can be generated by these methods, they often change the identity irrelevant attributes and bring extra visual artifacts. 
Reversible face de-identification are studied by a few works recently. \cite{gu2020password} performs anonymization and deanonymization with preset binary passwords. \cite{cao2021personalized} decouples a face image into attribute representation and identity representation, and applies vector rotation on the identity part to achieve reversible de-identification. The flexiblity and security of de-identification can be affected, since these approaches are usually based on complex designed encryption rules, or requires additional conditional information and hold-out face image set. Differently, our method uses randomly sampled latents as passwords, and can be trained in an end-to-end manner. 

\subsection{GAN Inversion}
GAN Inversion methods focus on embedding an existing image into the latent space of a pretrained generator. StyleGAN\cite{karras2019style,karras2020analyzing} is the most commonly used generator architecture for inversion, which can generate high resolution and photo-realistic images with multi-level latent codes as its inputs. Optimization based methods \cite{abdal2019image2stylegan,abdal2020image2stylegan++,karras2020analyzing,zhu2020domain} regard latent codes as trainable parameters and iteratively update their value by minimizing several reconstruction loss terms. These methods can acquire plausible reconstruction results, but can also suffer from poor edit-ability. 
Encoder based methods \cite{tov2021designing,richardson2021encoding,alaluf2021restyle}  tend to learn a general mapping from images to the latent manifold, which are much faster with a few sacrifices for reconstruction results. 
In this work, we use one typical encoder-based inversion method  \cite{tov2021designing} to extract latent codes from the face images as our training data. 

\section{Method}

In this section, we detail our \method framework. During the encryption phase, the latent code and the password are processed by a lightweight latent encryptor, which has a transformer-based architecture, to yield the encrypted code. During decryption, the same encryptor takes the encrypted code and the password as input, and returns the decrypted code. To achieve diversified identity generation, an identity diversity loss term is imposed on the encryptor to separate all the de-identified faces from each other. To recover the original identity, the similarity between the correctly decrypted face and the original face is maximized. 
The full pipeline of \method is shown in Figure \ref{fig2}.


\subsection{Network Architecture}
The full model consists of a StyleGAN inversion encoder that maps the original image to the latent space, a latent encryptor that performs encryption and decryption, and a StyleGAN generator for final image generation. In the training process, we update the parameters of the encryptor while keeping the other components fixed.

Our latent encryptor has a transformer-based  architecture, and combines the latent code with the password via cross-attention. 
Firstly, the latent code to be encrypted $\mathbf{w}$
is chunked into three slices to get $\mathbf{w}^c$, $\mathbf{w}^m$ and $\mathbf{w}^f$ , corresponding to the different levels (coarse, medium and fine) of the StyleGAN generator, respectively. The same operation is imposed on the password, which is a randomly sampled latent. 
Then, the latent code slice and the password slice at the same level are processed by their corresponding transformer block. 
The multi-head attention mechanism makes the code condition on its password, and help the encryptor to learn identity sensitive information. 
Finally, the output of each transformer block is concatenated together and passed through a fully connected layer to form the encrypted code. We can yield the de-identified face by feeding the encrypted code into the StyleGAN generator. The decryption process is similar, with only the original code replaced with the encrypted one.

\subsection{Latent Encryption and Decryption}

In the training phase of the encryptor, each training step consists of two forward propagations, one for encryption and another for decryption. For an original face image $\mathbf{x}$, its latent code $\mathbf{w}$ together with $m$ randomly sampled passwords are fed into the encryptor, and we can get the encrypted codes. Then, a second forward pass through the same encryptor is performed, where the inputs are the encrypted codes and the passwords for decryption. Each encrypted code is decrypted by another $n$ randomly sampled wrong passwords and its own correct password, resulting in $n+1$ decrypted codes. Losses are calculated and backpropagated after the second forward pass. 



\textbf{Diverse Identities Generation.}  
Previous de-identification methods tend to directly push the anonymous identity away from the original one, while optimizing a pixel level reconstruction loss to maintain the overall structure. Under such constraints, it is easy for the model to generate all anonymous faces with similar appearances. 
Some methods achieve diversity by pushing the generated face close to a guide at feature level through multiple discriminators or loss functions. The guide images are usually selected from a hold-out identity set or directly come from the training set.  
In this case, the diversity is limited to a large extent and some attributes of the guide will leak into the anonymous face,  resulting in unnatural visual artifacts. 
Differently, we use an identity diversity loss term to minimize the identity feature similarity between each pair of the de-identified images, thereby boost diversity.


Taking a single image as an example, there exist $m$ passwords for identity encryption and each password has another $n$ wrong  passwords for decryption, and all $m \times (n+1)$ passwords used for encryption and incorrect decryption are unique. We first use a pretrained Arcface network \cite{deng2019arcface} $F_{e}$ to extract the identity embeddings of all the faces. Then, the similarity of the embeddings of all the encrypted and wrongly decrypted faces can form a square matrix $\mathcal{M}$ with $m \times (n+1)$ rows. The arbitrary element at row $i$ and column $j$ can be given by
\begin{equation}
\mathcal{M}_{ij}=\max \left(\epsilon, \cos \left(F_e(\mathbf{x}^i), F_e(\mathbf{x}^j)\right)\right),
\end{equation}
where $\cos$ denotes for cosine similarity, and $\epsilon$ represents a predefined similarity threshold which is set to 0 in all the experiments.   $\mathbf{x}^i$ can be an encrypted image
or a wrongly decrypted image, so as $\mathbf{x}^j$.
Our identity diversity loss term can be written as 
\begin{equation}
    \mathcal{L}_{div}=\frac{1}{m^2 {\left(n+1 \right)}^2} \cdot  \mathbf{sum} (\left(\mathbf{1}-\mathbf{I}\right) \cdot \mathcal{M}),
\end{equation}

where $\mathbf{1}$ is a matrix with all ones and $\mathbf{I}$ is the unit diagonal matrix, and $\cdot$ denotes for elementwise multiplication. By minimizing the sum of cosine similarity between each pair of de-identified images, the loss associates each password with a unique identity.

\textbf{Reversible De-identification.}
To achieve the goal of identity encryption, we simply push the de-identified faces away from the original face $\mathbf{x}$ in the identity embedding space. For $m$ encrypted faces and $m \times n$ wrongly decrypted faces, the similarity between them and the original face need to be minimized. Thus, the de-identification loss can be written as 
\begin{equation}
\mathcal{L}_{deid}=\sum_{i}^{m (n+1)} \max \left(\epsilon, \cos \left(F_e(\mathbf{x}), F_e(\mathbf{x}^i)\right)\right).
\end{equation}
If the password is correct, then the decrypted latent code should be able to recover the true identity, and the similarity between the generated face and the original face should be maximized. Thus, the identity recovery loss term can be written as 
\begin{equation}
\mathcal{L}_{rec}=\sum_{i}^{m} \left(1 - \cos \left(F_e(\mathbf{x}), F_e(\mathbf{x}^{i}_{corr})\right)\right),
\end{equation}
where $\mathbf{x}^{i}_{corr}$ is the image correctly decrypted by $i$th password.

Our full identity loss term can be written as
\begin{equation}
\mathcal{L}_{id}=\mathcal{L}_{div}+\mathcal{L}_{deid}+\mathcal{L}_{rec}.
\end{equation}

\textbf{Face Quality Assurance.}
We utilize a pixelwise reconstruction loss to assist the  decryption process meanwhile keeping the overall similarity. This loss term is calculated between all the generated images (including the encrypted ones and the correctly and wrongly decrypted ones, denoted as $\mathbf{x^*}$) and the original one, which can be written as   

\begin{equation}
\mathcal{L}_{pix}=\left\| \mathbf{x} - \mathbf{x}^{*}\right\|_{1}.
\end{equation}

Besides, we apply a LPIPS loss term \cite{zhang2018unreasonable} on all the generated images, for better preserving image quality and similarity at the feature level, formulated as 

\begin{equation}
\mathcal{L}_{LPIPS }=\|F_{p}(\mathbf{x})-F_{p}(\mathbf{x}^{*})\|_{2},
\end{equation}
where $F_p$ stands for the pretrained perceptual feature extractor. 

During training, we observed that sometimes the expression of the generated image will change together with the shift of identity, and in a few cases, the shape of some facial features will become unnatural, e.g. overlarge nose, eyes or mouth. We use a face parsing loss

\begin{equation}
\mathcal{L}_{parse}=\|F_{s}(\mathbf{x})-F_{s}(\mathbf{x}^{*})\|_{2},
\end{equation}
to avoid these artifacts, where $F_s$ represents an off-the-shelf face parsing model from CelebAMask-HQ \cite{CelebAMask-HQ}. We only select the channels corresponding to the eyes, ears, mouth, and nose while calculating the loss. 

To keep the generated latent codes staying in the well defined manifold of StyleGAN, we apply a latent regularization loss on all the encrypted and decrypted codes (denoted as $\mathbf{w}^{*}$), the latent regularization loss is given by
\begin{equation}
\mathcal{L}_{latent}=\left\| \mathbf{w} - \mathbf{w}^{*}\right\|_{2}.
\end{equation}

Our final loss function can be formulated as
\begin{equation}
\begin{aligned}
\mathcal{L}_{total}= & \mathcal{L}_{id}+\lambda_{pix}\mathcal{L}_{pix}+\lambda_{LPIPS}\mathcal{L}_{LPIPS}+ \\  & \lambda_{parse}\mathcal{L}_{parse}+\lambda_{latent}\mathcal{L}_{latent}.
\end{aligned}
\end{equation}
where $\lambda_{pix} $, $\lambda_{LPIPS}$  $\lambda_{parse}$ and $\lambda_{latent}$ are hyperparameters which control the magnitude of different losses. We set $\lambda_{{pix}}$ to 0.05, $\lambda_{{LPIPS}}$ to 1, $\lambda_{{parse}}$ to 0.1 and $\lambda_{latent}$ to 0.01.  

\subsection{Data-free Training} 
When the training dataset is unaccessible, the latent encryptor can be trained in a data-free setting. Specifically, instead of the inverted latents from real images, we can use the randomly sampled codes in the latent space of the StyleGAN to train the encryptor. Although there exists some degradation in image quality, the de-identification performance is almost unaffected. Performance of the model trained in the data-free setting are shown in the experiments part. 


\section{Experiments}

\begin{figure*}[t]
\centering
\includegraphics[width=0.8\linewidth]{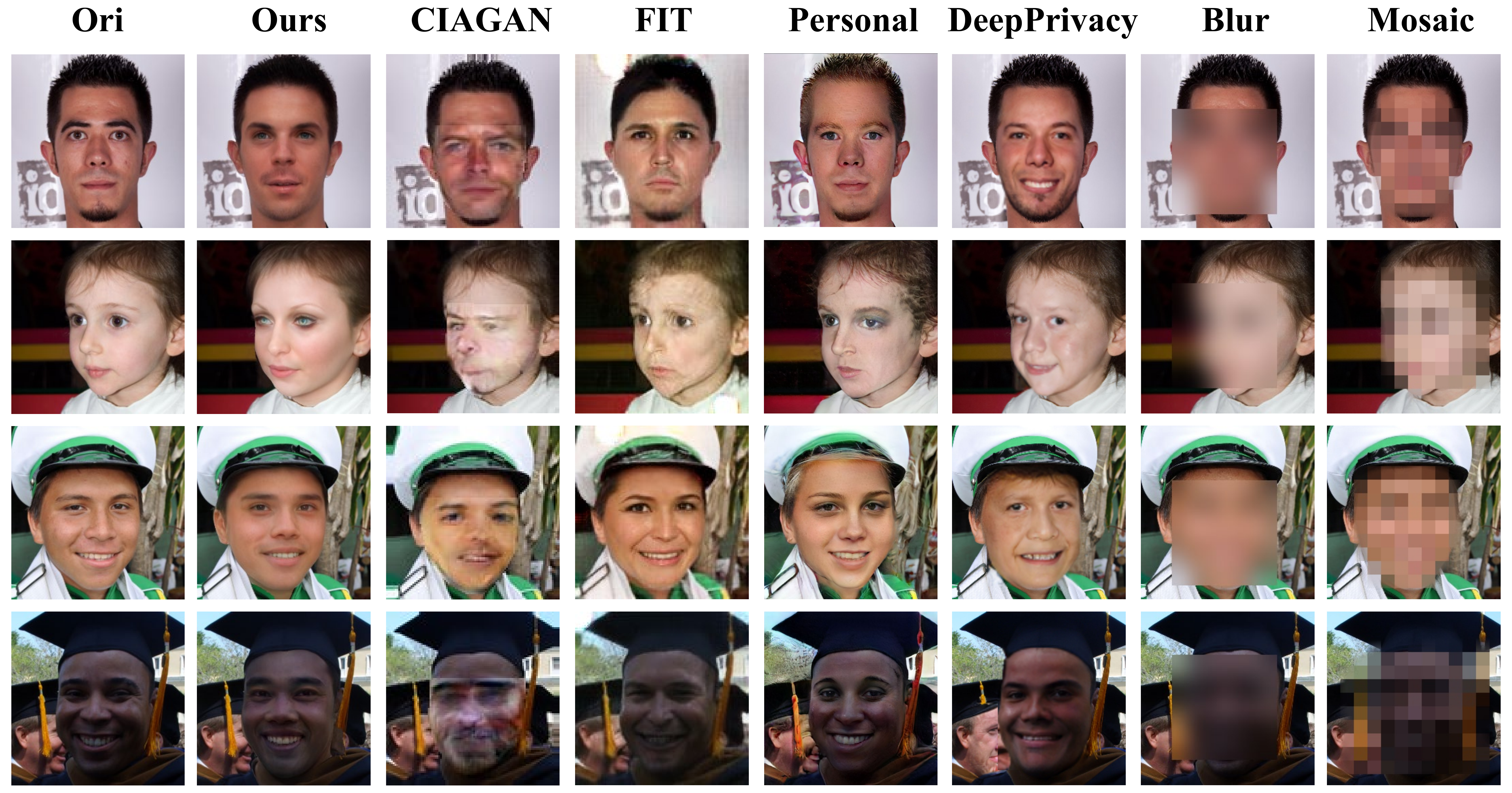} 
\caption{Qualitative comparison with literature de-identification methods, FFHQ. \method has the best image quality, meanwhile keeping the identity irrelevant attributes (e.g. pose, texture,  expressions) well.}
\label{fig3}
\end{figure*}

\subsection{Experiment Setup}
\textbf{Basic Settings.} The number of passwords used for encryption and decryption, i.e. $n$ and $m$, are both set to 2. We train our model on 4 NVIDIA TITAN RTX GPUs and the total batchsize is set to 8. The whole training process requires 200,000 steps and lasts approximately  two days. 
More training details can be found in the supplementary material. 

\textbf{Datasets.} We use FFHQ \cite{karras2019style} for training and demonstrating the results. CelebA-HQ dataset \cite{karras2017progressive} is used for evaluating the image quality and utility for downstream tasks. For calculating identification rates for de-identification and recovery performance, we use the Labeled Faces in the Wild \cite{LFWTech}  dataset, following a standard test protocol for face verification. We also apply our model to images and video sequences in the wild to show the generalization ability of our model.

\begin{table}[t]
\scalebox{0.85}{

\begin{tabular}{@{}ccccc@{}}
\toprule
Type                               & Method       & \thead{FaceNet\\CASIA} & \thead{FaceNet\\VGGFace2} & SphereFace \\ \midrule
\multirow{5}{*}{De-id $\downarrow$} & Ours        & \textbf{0.016}       & \textbf{0.032}            & 0.025      \\
                                    & Ours-DF        & 0.034      & 0.037            & 0.025   \\
                                    & CIAGAN \cite{maximov2020ciagan}    & 0.019         & 0.034            & \textbf{0.010}       \\
                                   & FIT \cite{gu2020password}         & 0.042         & 0.072            & 0.065      \\
                                   & Personal \cite{cao2021personalized}        & 0.020        & 0.042            & 0.017   \\
                                   & DeepPrivacy\cite{hukkelaas2019deepprivacy} & 0.266         & 0.184            & 0.120       \\ \midrule
\multirow{2}{*}{Recovery $\uparrow$}          & Ours        & \textbf{0.996}         & \textbf{0.998}            & \textbf{1.000}          \\
& Ours-DF        & 0.953         & 0.949            & 1.000 \\
                                   & FIT  \cite{gu2020password}      & 0.967         & 0.974            & \textbf{1.000 }         \\
                                   & Personal \cite{cao2021personalized}        & 0.965        & 0.965       & 0.998  \\
                                   \bottomrule
\end{tabular}
}

\caption{Identification rate comparison. Lower identification rate implies better anonymization, while higher identification represents beter recovery. Our method achieves competitive performance. }
\label{tab2}
\end{table}

\subsection{De-identification}
We compare \method with two conditional-GAN based methods: CIAGAN \cite{maximov2020ciagan} and Face Identity Transformer (shorten as FIT) \cite{gu2020password}, one face-inpainting based method DeepPrivacy \cite{hukkelaas2019deepprivacy} and an identity disentanglement based method Personal \cite{cao2021personalized}. Face blurring and mosaicing are also taken into consideration. 

A qualitative comparison can be seen in Figure \ref{fig3}. Blurring and Mosaicing can successfully erase the identity information, but the photo-realism has been damaged severely. FIT\cite{gu2020password} can generate diversified faces, however, their visual quality is not satisfactory and the generated face texture is affected. CIAGAN \cite{maximov2020ciagan} tends to generate seriously distorted images with poor diversity. DeepPrivacy \cite{hukkelaas2019deepprivacy} can maintain photo realism to some extent but the generated faces have similar appearance. It also fails to retain the identity-irrelevant attributes such as expressions. Besides, in some cases, the generated face is not aligned with its contour. Peronsal \cite{cao2021personalized} brings additional artifacts such as wrinkles, different hair color, beards and so on, which is caused by the 
attribute leakage from the guide image. 
Different from the above methods, \method generates smoother images with better photo-realism. What's more, attributes such as pose and expressions are held at the same time.


Quantitatively, we calculate the identification rate on the LFW dataset. We only compare our method with  deep learning based methods. Lower identification rate indicates better anonymization. Our model trained in the data-free setting is denoted as Our-DF. The results are shown at the top of Table \ref{tab2}. Notice that we use different face recognition backbones from which we used for calculating identity losses during training. This can further validate the generalization ability of our method. It can be seen that our method and \cite{maximov2020ciagan} surpass other methods by a large margin. Moreover, our method can generate more realistic faces than the latter. 

\begin{figure}[t]
\centering
\includegraphics[width=0.9\linewidth]{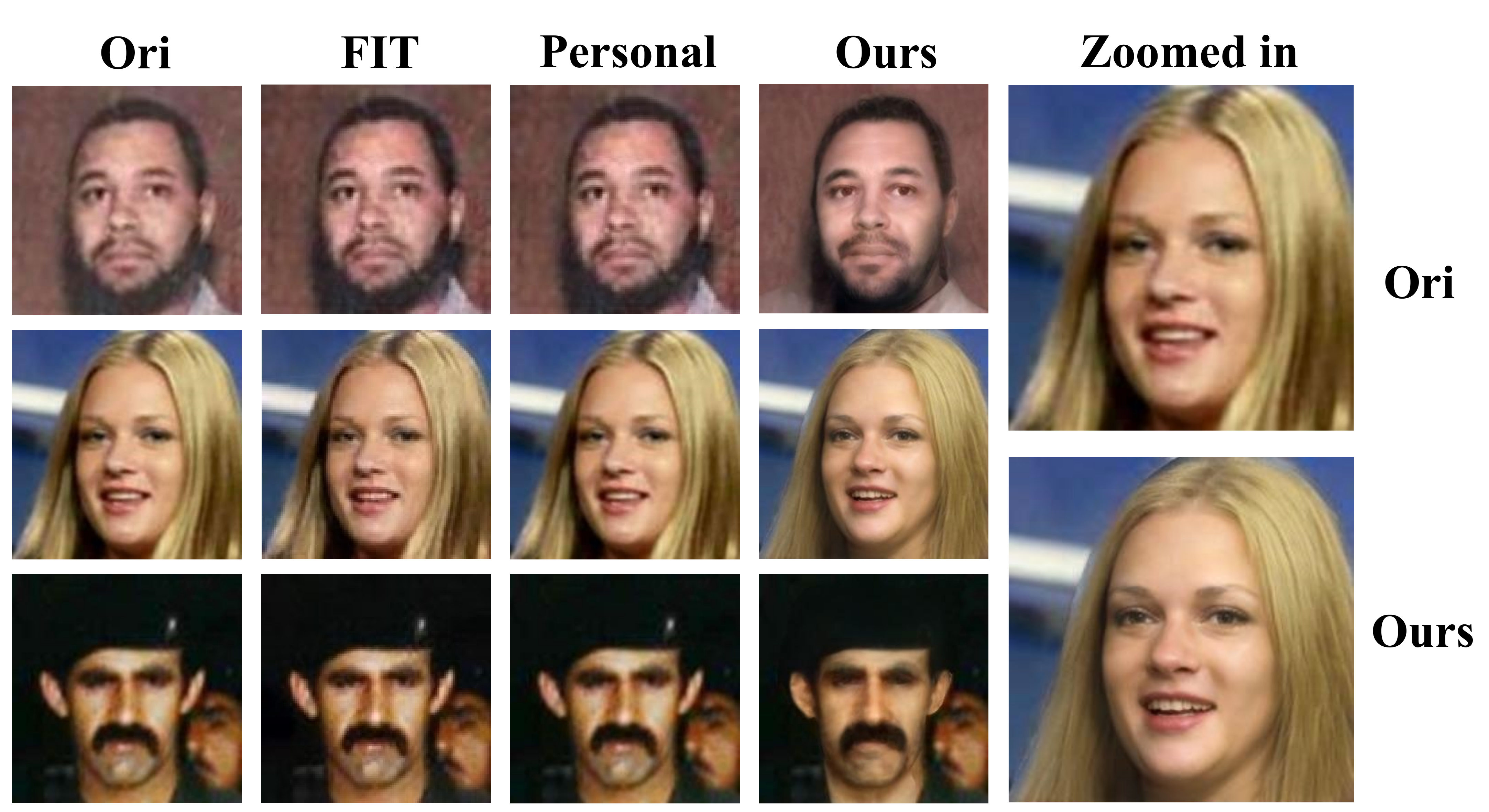} 
\caption{Demonstration of identity recovery, LFW. Our recovered image can even be regarded as the super-resolution version of the original one.}
\label{fig_rec}
\end{figure}

\begin{table}[t]
\centering
\scalebox{0.85}{
\begin{tabular}{@{}ccccc@{}}
\toprule
      & MSE$\downarrow$ & LPIPS$\downarrow$ & SSIM$\uparrow$ & PSNR$\uparrow$ \\ \midrule
FIT \cite{gu2020password}    &  0.005   &     0.186   &    0.934   &   23.130   \\
Personal \cite{cao2021personalized} &  0.003   &   0.220    &    0.846   &   26.391   \\ 
Ours   &  \textbf{0.002}   &     \textbf{0.043}   &    \textbf{0.966}   &   \textbf{26.499} \\
Ours-DF   &  0.004   &     0.277   &    0.760   &   25.483 
\\ \bottomrule
\end{tabular}
}
\caption{Image quality metric comparison on recovered images, Celeb-HQ and FFHQ. }
\label{tab3}
\end{table}

\subsection{Identity Recovery}
For identity recovery, we compare our method with FIT \cite{gu2020password} and Personalized \cite{cao2021personalized}. Results are shown in the bottom part of Table \ref{tab2}. Higher identification rate represents better recovery. Then, we calculate the similarity between the correctly decrypted images and the original images.
 We use the mean square error, PSNR (peak signal-to-noise ratio), SSIM (structural similarity), and LPIPS \cite{zhang2018unreasonable} as our metrics. Results are shown in Table \ref{tab3}. Identity recovery on LFW datasets is more challenging due to the lower resolution and the complex surroundings of each individual. However, leveraging the strong generative ability of StyleGAN2, \method can successfully restore the identity. As is shown in Figure \ref{fig_rec}, images recovered by our method are sharper, smoother, and have fewer artifacts than \cite{gu2020password} and \cite{cao2021personalized}. Some high-frequency details are added by our model during the recovery process and our recovery process can be considered to be similar to super-resolution.

\begin{figure}[t]
\centering
\includegraphics[width=0.85\linewidth]{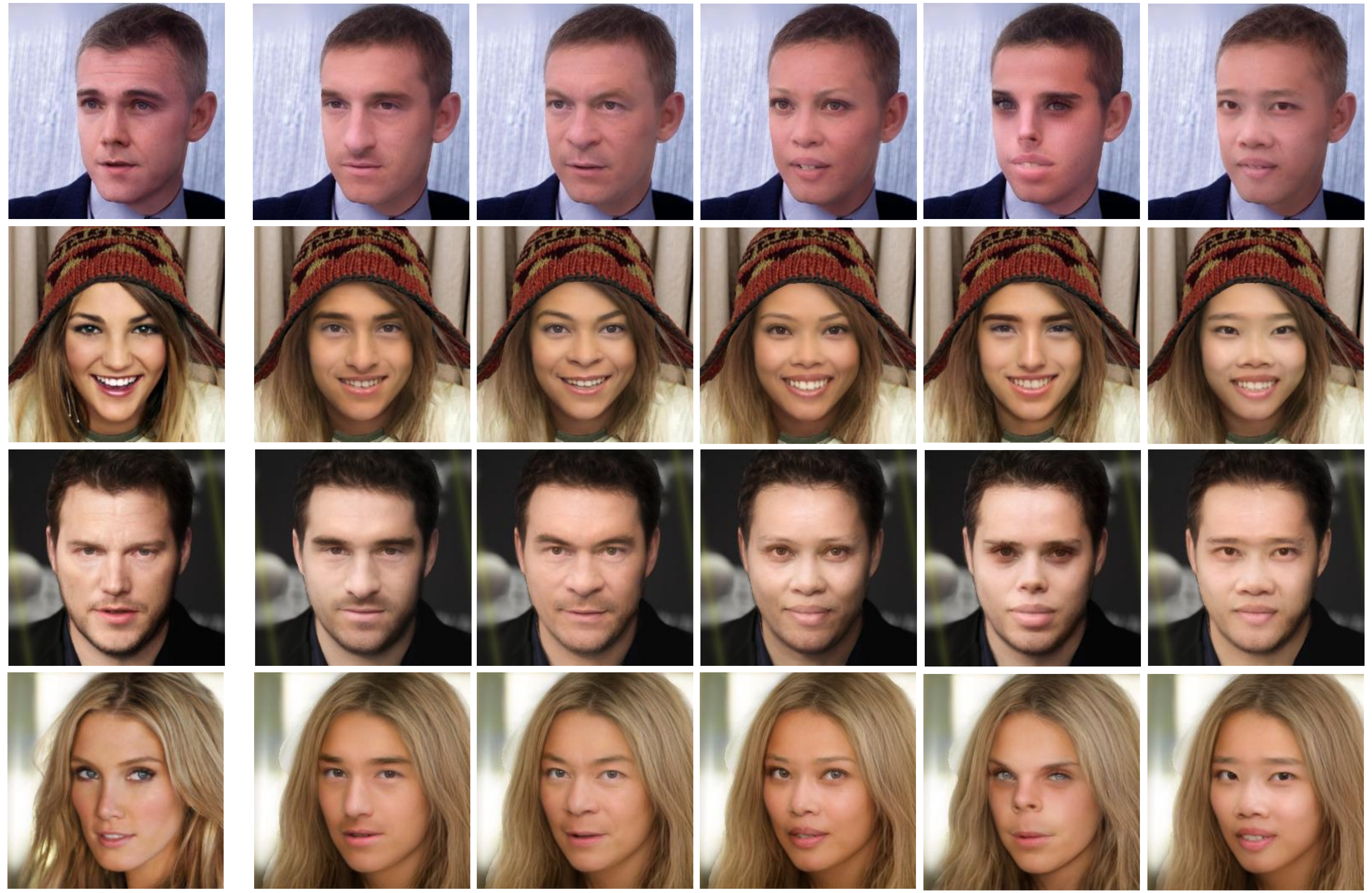} 
\caption{Various de-identification results conditioned on different passwords, CelebA-HQ. The leftmost column is the original images, the rest columns are de-identified images with different passwords. Images in the same column share the same password.}
\label{fig4}
\end{figure}

\subsection{Diversity of Identities}


\begin{table*}[t]
\centering
\scalebox{0.9}{
\begin{tabular}{@{}cccccccc@{}}
\toprule
\multicolumn{2}{c}{Method} &
  Ours &
  Ours-DF &
  CIAGAN \cite{maximov2020ciagan} &
  FIT \cite{gu2020password} &
  Personal \cite{cao2021personalized} &
  DeepPrivacy \cite{hukkelaas2019deepprivacy} \\ \midrule
\multicolumn{2}{c|}{FID $\downarrow$} &
  \textbf{15.389} &
  26.802 &
  32.611 &
  30.331 &
  25.715 &
  23.713 \\ \midrule
\multicolumn{1}{c|}{\multirow{2}{*}{\thead{Face detection $\uparrow$}}} &
  \multicolumn{1}{c|}{MtCNN} &
  \textbf{1.000} &
  1.000 &
  0.992 &
  \textbf{1.000} &
  \textbf{1.000} &
  \textbf{1.000} \\
\multicolumn{1}{c|}{} &
  \multicolumn{1}{c|}{Dlib} &
  0.991 &
  0.975 &
  0.937 &
  0.984 &
  \textbf{0.992} &
  0.980 \\ \midrule
\multicolumn{1}{c|}{\multirow{2}{*}{\thead{Bounding box \\ distance $\downarrow$}}} &
  \multicolumn{1}{c|}{MtCNN} &
  \textbf{3.824} &
  5.720 &
  20.387 &
  7.879 &
  4.213 &
  4.654 \\
\multicolumn{1}{c|}{} &
  \multicolumn{1}{c|}{Dlib} &
  \textbf{1.700} &
  3.109 &
  15.476 &
  4.218 &
  2.726 &
  2.685 \\ \midrule
\multicolumn{1}{c|}{\multirow{2}{*}{\thead{Landmark \\distance $\downarrow$}}} &
  \multicolumn{1}{c|}{MtCNN} &
  \textbf{1.674} &
  3.252 &
  8.042 &
  3.572 &
  2.358 &
  3.280 \\
\multicolumn{1}{c|}{} &
  \multicolumn{1}{c|}{Dlib} &
  \textbf{1.512} &
  2.973 &
  8.930 &
  4.047 &
  2.459 &
  2.896 \\ \bottomrule
\end{tabular}
}
\caption{Utility evaluation of de-identification results, CelebA-HQ.}
\label{tab4}
\end{table*}

\begin{figure}[t]
\centering
\includegraphics[width=0.9\linewidth]{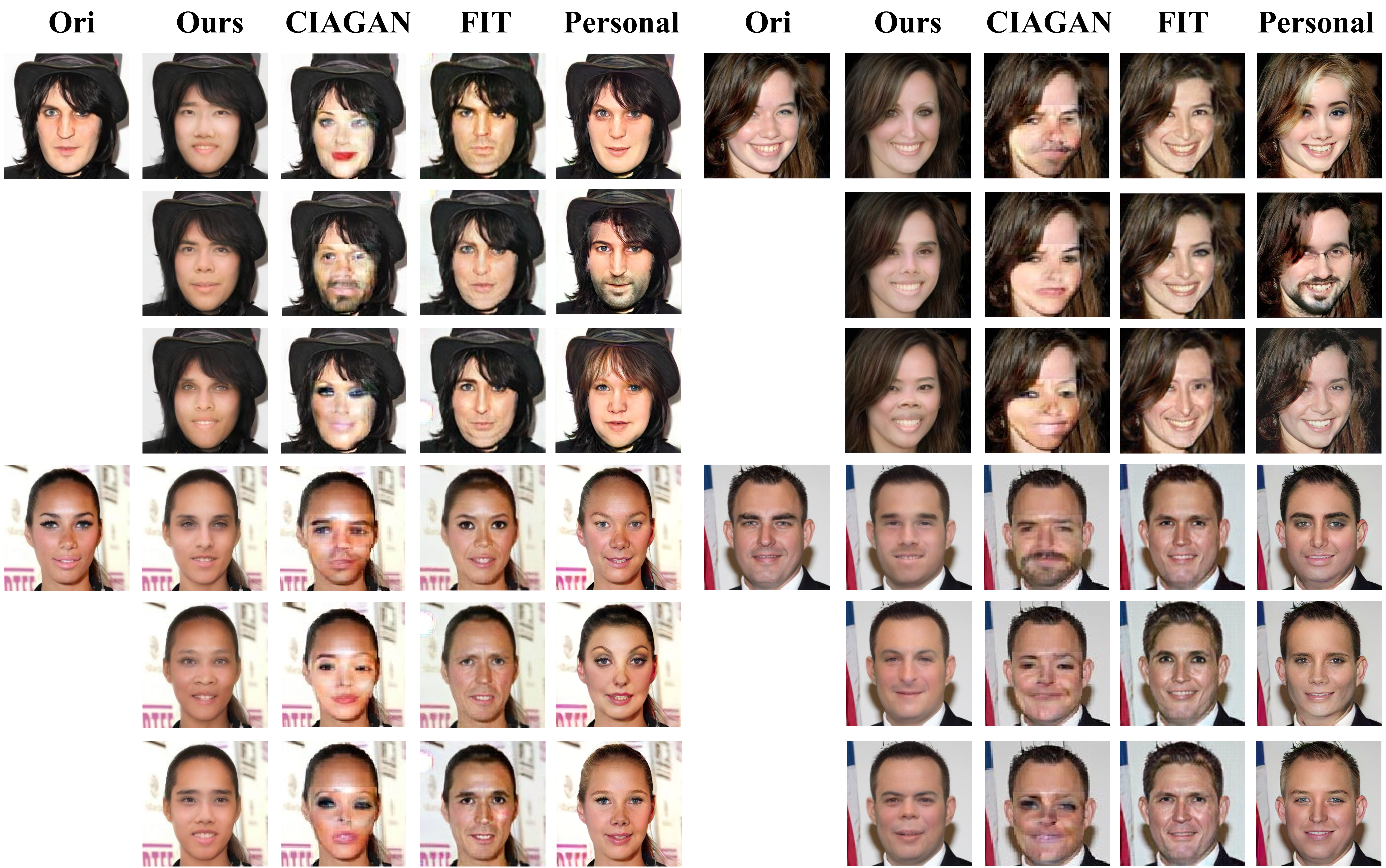} 
\caption{Quality comparison with literature methods, CelebA-HQ and FFHQ. For each identity, the first column is the original images, the second column is our result, and the rest columns are de-identified images generated by FIT\cite{gu2020password}, CIAGAN\cite{maximov2020ciagan}, and Peronal\cite{cao2021personalized}, respectively.}
\label{fig_div2}
\end{figure}

Figure \ref{fig4} shows the diversified de-identification results yielded by \method. The diversity of faces is demonstrated in the global and local faical features, such as the texture of the skin and the shape of the eyes and nose.
We compare our method with FIT \cite{gu2020password}, CIAGAN \cite{maximov2020ciagan}, and Personal \cite{cao2021personalized}, results are shown in Figure \ref{fig_div2}. Although all these methods guarantee diversity,  \cite{gu2020password} tends to introduce some common features in local regions for faces conditioned on different passwords.\cite{maximov2020ciagan} brings obvious splicing traces and its de-identified faces suffer from texture inconsistency. \cite{cao2021personalized} achieves diversity by changing identity irrelevant attributes. 
In contrast, our method brings fruitful facial features for different images, while keeping the other attributes well.

\begin{figure}[t]
\centering
\includegraphics[width=0.9\linewidth]{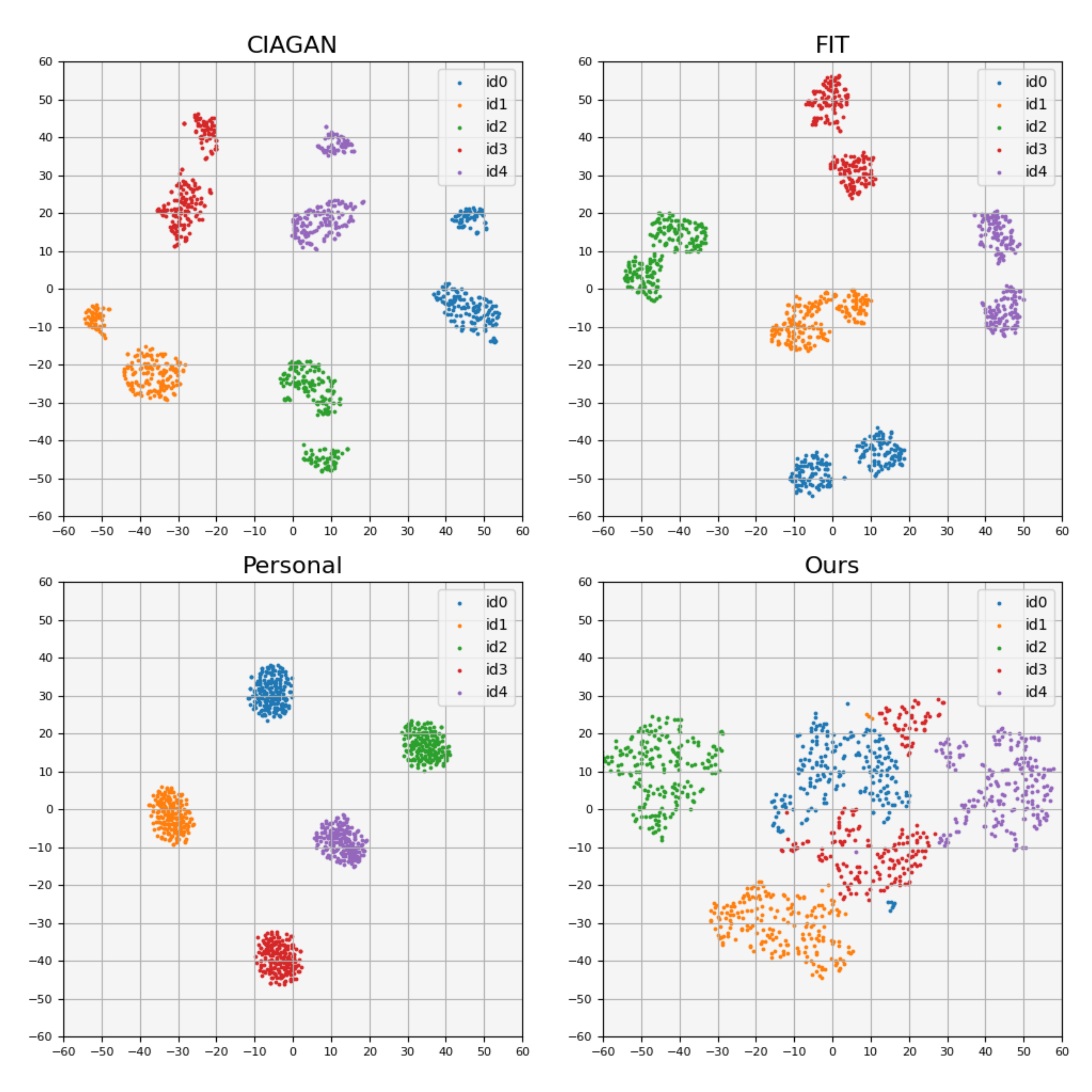} 
\caption{Quantitative comparison with literature methods on diversity. Diversity of identities generated by \method shows superiority over other methods.}
\label{fig_div3}
\end{figure}

To further demonstrate the diversity of our method, an identity visualization experiment  is conducted. First, we randomly sampled five people in CelebA-HQ dataset. For each person, we use 200 different passwords or labels to generate their de-identified faces. Then, we use an off-the-shelf ArcFace model to extract identity embeddings of all the faces and perform dimensionality reduction using t-sne\cite{van2008visualizing}. The distribution of identity embeddings generated with different methods is shown in Figure \ref{fig_div3}. It can be seen that our de-identified faces are more scattered and occupy most of the area of the hyperplane. Notice that the overlap of different clusters will disappear on the original high dimensional identity embedding space (512-D).  For other methods, embeddings from the same person are more likely to cluster, which indicates that despite successful anonymity, their diversity is limited. The identity cluster of Personal\cite{cao2021personalized} is the tightest because of the hold-out identity set and the direct feature mapping. 

We also calculated  Known Factor Feature Angle (KFFA) \cite{li2022contrastive} of our encrypted faces and the wrongly decrypted faces as a quantitative index for diversity evaluation, as shown in Table \ref{tabKFFA}. Curricularface \cite{huang2020curricularface} is used as the feature extractor and 10 de-identified faces are generated for each identity. It can be seen that RiDDLE has better diversity than other methods. Besides, the identity similarity between the de-identified face generated by RiDDLE and the original face is also lower compared to other methods. This demonstrates the ability of our approach to ensure diversity while maintaining security.

\subsection{Face Utility}

\textbf{Utility on downstream tasks.} We evaluate the utility of our de-identified images for downstream vision tasks. We calculate the face-detection rate of \method and literature methods on two different face detection models: MtCNN\cite{zhang2016joint} and Dlib \cite{kazemi2014one}. The per pixel distance of landmarks corresponding to different face areas are also calculated. The results are shown in Table \ref{tab4}. Although our de-identified images have a large variety of important facial features, the average pixel difference with the original face is relatively low, which means the overall facial information is better retained by our method. Lastly, FID of our method is significantly lower than other methods, which validates the quality and diversity of our de-identified faces.

\begin{table}[t]
\centering
\scalebox{0.65}{
\begin{tabular}{@{}c|cc|cc|cc@{}}
\toprule
Dataset           & \multicolumn{2}{c|}{FFHQ} & \multicolumn{2}{c|}{CelebA-HQ} & \multicolumn{2}{c}{LFW} \\ \midrule
Metric            & Id-Sim $\downarrow$      & KFFA $\uparrow$      & Id-Sim $\downarrow$        & KFFA   $\uparrow$       & Id-Sim $\downarrow$     & KFFA $\uparrow$      \\ \midrule
CIAGAN \cite{maximov2020ciagan}           & 0.039            &  64.29      & 0.068           & 62.14            & 0.050      & 62.58         \\
FIT \cite{gu2020password}            & 0.104            & 66.45      & 0.147           & 65.20            & 0.125           & 64.19         \\
Personal \cite{cao2021personalized}         & 0.192            & 54.71          & 0.187              & 58.29            & 0.205           & 56.53         \\
Encrypted         & \textbf{0.038}        & \textbf{74.88}      & 0.056         & \textbf{77.39}        & 0.058       & \textbf{75.83}     \\
Wrongly Decrypted & 0.052        & 73.07      & \textbf{0.054}          & 74.04        &\textbf{0.028}       & 73.47     \\ \bottomrule
\end{tabular}

}
\caption{Mean identity cosine similarity with the original faces and KFFA metric. Both the encrypted faces and the wrongly decrypted faces have low cosine similarity with the original images, while maintaining a high KFFA value.}
\label{tabKFFA}
\end{table}

\begin{figure}[t]
\centering
\includegraphics[width=0.9\linewidth]{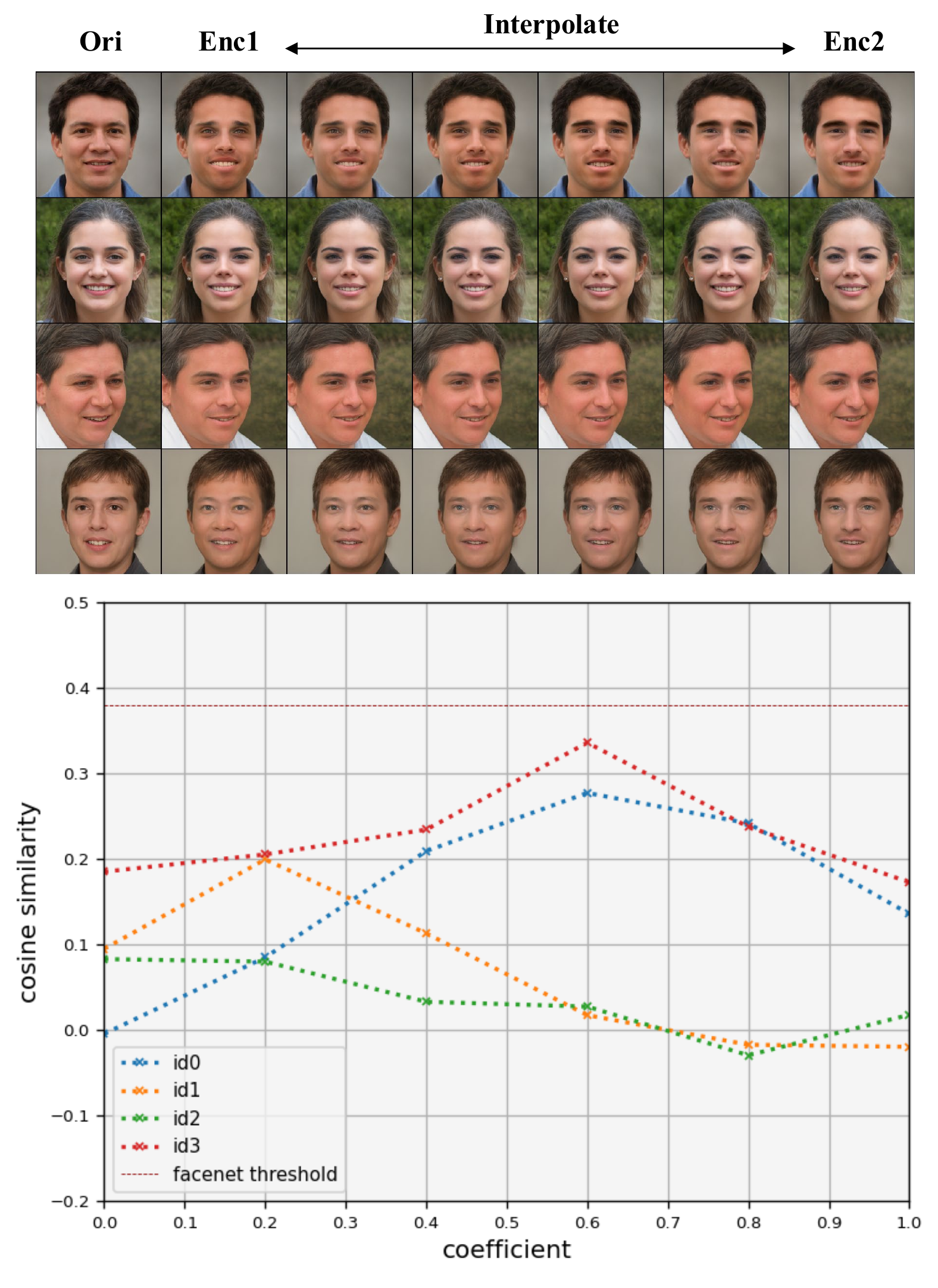} 
\caption{Password interpolation results. Upper: Encrypted faces.
Lower: Identity distance between the original and the encrypted faces obtained by password interpolation.}
\label{fig_int}
\end{figure}

\textbf{Password Interpolation.} To further study the property of our password scheme, we perform interpolation between two passwords, and use the intermediate results to encrypt and generate images. As Figure \ref{fig_int} shows, while performing interpolation between two passwords for encryption, the identities of the generated images have shown a smooth transition, while other attributes remain fixed.
This further validates the ability of RiDDLE on attribute retention and keeping image quality. Each password is linked to a unique identity, while the similarity between the original image and the encrypted faces is lower than the face recognition threshold without exception.


\textbf{Video De-identification.} It is noteworthy that \method can be applied to video sequences. Specifically, we first detect the face in each video frame and perform de-identification, then we paste the processed faces back to their original position. High-quality de-identified videos can be seen in the supplementary materials. 


\subsection{Ablation Study}
\begin{figure}[t]
\centering
\includegraphics[width=0.95\linewidth]{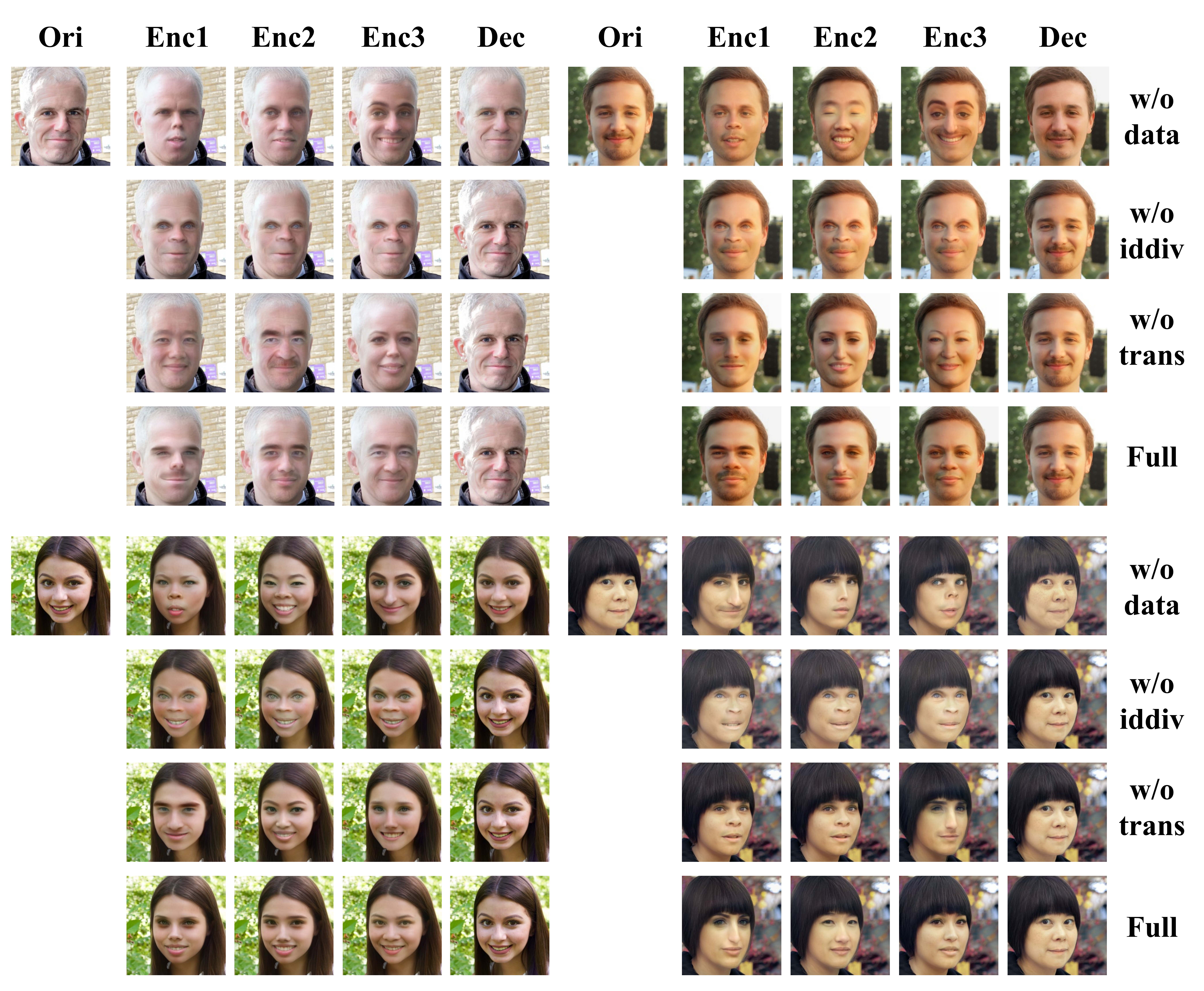} 
\caption{Qualitative ablation study on some important components of RiDDLE. Results on four different identities are shown. }
\label{fig_ablation}
\end{figure}

The function of some key components of RiDDLE are analyzed in this part. 
\textit{w/o trans} indicates using a simple encryptor that consists of three fully connected layers with leaky-relu activations as a substitute. \textit{w/o iddiv} represents a model trained without the identity diversity loss term and \textit{w/o data} denote a model obtained under the data free setting.
Qualitative and quantitative results are shown in Figure \ref{fig_ablation} and Table \ref{tab_ablation} respectively. 
It can be seen that the transformer structure can greatly improve the image quality, thus brings better de-identification and restoration. The model obtained from data-free training holds a competitive de-identification performance, but suffers from lower quality. Moreover, the ability of maintaining the identity independent attributes (such as expression) has been affected. Without the identity diversity loss term, the model can only generate images with the same identity. However, the identity recovery ability is nearly unaffected.

\begin{table}[t]
\centering
\scalebox{0.8}{
\begin{tabular}{@{}cccc@{}}
\toprule
                            & De-id $\downarrow$ & Recovery $\uparrow$ & FID $\downarrow$ \\ \midrule
w/o data                    & 0.034             & 0.953          & 26.802           \\
w/o transformer             & 0.018             & 0.985          & 22.704           \\
w/o identity diversity loss & 0.025             & 0.993          & 25.816           \\
full                        & \textbf{0.016}             & \textbf{0.996}          & \textbf{15.389 }          \\ \bottomrule
\end{tabular}
}
\caption{Quantitative ablation study. Identification rate of de-identification and recovery, and FID are calculated.}
\label{tab_ablation}
\end{table}



\section{Conclusion}
This paper proposes \method, which is short for \textbf{R}evers\textbf{i}ble and \textbf{D}iversified \textbf{D}e-identification with \textbf{L}atent \textbf{E}ncryptor. \method performs encryption and decryption on the latent space, and can generate various de-identified faces with high fidelity. Compared with literature methods, \method achieves better quality and diversity and is data-efficient. \method can be used for various privacy-sensitive scenarios such as video conferences and security monitoring. 
In the future, we are about to explore more intriguing properties of the identity manifold. 

\section*{Acknowledgements}
This work was supported by the National Key Research and Development Program of China under Grant
No. 2020AAA0140003 and the National Natural Science Foundation of China (NSFC) under Grant 61972395, and by Alibaba Group through Alibaba Innovative Research Program. We would like to thank Yujun Shen for his advice. 

{\small
\bibliographystyle{ieee_fullname}
\bibliography{egbib}
}

\end{document}